\documentclass[journal]{IEEEtran}
\pdfoutput=1
\usepackage{graphicx}
\usepackage[colorlinks,
            linkcolor=red,
            anchorcolor=green,
            citecolor=green]{hyperref}
\usepackage{amsmath}
\usepackage{caption}
\usepackage{array}
\usepackage{booktabs}
\usepackage{longtable}
\usepackage{tabularx, makecell, multirow}
\usepackage{cite}
\usepackage{algorithm}  
\usepackage{algorithmic}
\usepackage{setspace}
\begin{document}
\title{Contour Loss for Instance Segmentation via \emph{k-step} Distance Transformation Image}
\author{Xiaolong Guo,
        Xiaosong Lan,
        Kunfeng Wang,
        Shuxiao Li$^{\dagger}$
\thanks{------------------------------------------------------------------------------------------}
\thanks{
Xiaolong Guo and Shuxiao Li are with the School of Artificial Intelligence, University of Chinese Academy of Sciences (UCAS) and Institute of Automation Chinese Academy of Sciences (CASIA) , Beijing, China (e-mail:$\{$guoxiaolong2018, shuxiao.li$\}$@ia.ac.cn).}
\thanks{Xiaosong Lan is with the Institute of Automation Chinese Academy of Sciences (CASIA), Beijing, China (e-mail: lanxiaosong2012@ia.ac.cn).}
\thanks{Kunfeng Wang is with the College of Information Science and Technology, Beijing University of Chemical Technology (BUCT) , Beijing, China (e-mail: wangkf@mail.buct.edu.cn).}
\thanks{This paper is a preprint of a paper submitted to IET Computer Vision. If accepted, the copy of record will be available at the IET Digital Library.}
\thanks{$^{\dagger}$Corresponding author.}}
\maketitle
\begin{abstract}
Instance segmentation aims to locate targets in the image and segment each target area at pixel level, which is one of the most important tasks in computer vision. Mask R-CNN is a classic method of instance segmentation, but we find that its predicted masks are unclear and inaccurate near contours. To cope with this problem, we draw on the idea of contour matching based on distance transformation image and propose a novel loss function, called contour loss. Contour loss is designed to specifically optimize the contour parts of the predicted masks, thus can assure more accurate instance segmentation. In order to make the proposed contour loss to be jointly trained under modern neural network frameworks, we design a differentiable \emph{k-step} distance transformation image calculation module, which can approximately compute truncated distance transformation images of the predicted mask and corresponding ground-truth mask online. The proposed contour loss can be integrated into existing instance segmentation methods such as Mask R-CNN, and combined with their original loss functions without modification of the inference network structures, thus has strong versatility. Experimental results on COCO show that contour loss is effective, which can further improve instance segmentation performances.
\end{abstract}
\section{Introduction}
\IEEEPARstart{I}{nstance} segmentation is a basic problem in computer vision. The main task of instance segmentation is to acquire the location and the pixel-wise semantic information of each instance. Benefited from tremendous development of deep learning \cite{p1} in object detection and semantic segmentation, instance segmentation based on deep learning has a rapid progress over a short period of time. However, due to diversity of objects and overlap between them, instance segmentation is still a challenging problem.

Taking classic instance segmentation method Mask R-CNN \cite{p2} as an example, although it can predict a general mask, contour of the mask (called \emph{predicted contour} later) is neither clear nor accurate. As far as we know, it could be a fatal problem in some applications. For example, in vision-based robot grabbing, clear and accurate contour is essential to the quality of grab detection. Our goal in this work is to make the predicted mask and its ground-truth mask not only consistent on the whole, but as consistent as possible near the contour.

Recently, some studies suggest that introducing additional supervisory signals beyond \emph{RGB} data may provide new clues in a complementary mode to improve performance of instance segmentation \cite{p3,p4,p5}. Those works respectively introduce depth, shape and key points as auxiliary information to efficiently perform instance segmentation. Although those methods achieve good performance from different perspectives, they usually require complicated pip-lines or large amount of training parameters. Moreover, the segmentation results near the contour are still not ideal.

To improve instance segmentation accuracy near the contour, this paper borrows classic idea of image distance transformation technology \cite{p6}, and brings distance transformation image (DTI) into instance segmentation for providing contour supervisory signals. Two main contributions are included in our method. \textbf{One aspect} is that we propose a novel loss function named contour loss based on DTI to optimize the contour part specially. In particular, we firstly calculate predicted \emph{k-step} DTI and ground-truth \emph{k-step} DTI for the predicted mask and the ground-truth mask respectively. Then, we accumulate coverage values of one contour image onto \emph{k-step} DTI of the other contour image, and the average of the two normalized coverage values is regarded as the difference measure between the predicted contour and the ground-truth contour. We define contour difference calculated in this manner as \emph{contour loss}, which can be integrated into existing instance segmentation algorithms such as Mask R-CNN without modifying their neural network structures. \textbf{The other aspect} is that we design a differentiable \emph{k-step} DTI calculation module, which approximately computes truncated DTIs of the predicted mask and the ground-truth mask online. The proposed module can be jointly trained in modern neural network frameworks without addition of other training parameters. To our best knowledge, this is the first analytic DTI module adaptive to current neural network frameworks. Experiments on COCO \cite{p7} show that the proposed contour loss is effective to produce more accurate and clearer masks, and can further improve the instance segmentation performances.

The rest of this paper is organized as follows: Section \ref{sec2} introduces related works. Section \ref{sec3} details the proposed contour loss. Section \ref{sec4} illustrates experiments to verify contour loss. Section \ref{sec5} draws main conclusions.
\section{Related Works}\label{sec2}
This section firstly introduces mainstream instance segmentation algorithms to date, and then summarizes separately those combining edge or boundary information where the differences between our method and theirs are analyzed.
\subsection{Mainstream Instance Segmentation Methods}\label{subsec2.1}
According to the processing pipe-line, current mainstream instance segmentation algorithms can be categorized into \emph{segmentation-based} methods and \emph{detection-based methods}.

\emph{Segmentation-based} methods firstly perform semantic segmentation in the image and then produce instance masks based on semantic information combination. FCN \cite{p8} has achieved remarkable success in the field of semantic segmentation, and numerous researchers have tried to apply it to instance segmentation. Dai et al. proposed Instance-FCN \cite{p9}. This method firstly generated a set of instance sensitive score maps which were used to predict semantic information of different relative positions of the same instance, and then got object masks through assembly accordingly. Li et al. proposed FCIS \cite{p10}. The authors used the feature representation of inside/outside position sensitive score maps to solve the problem that the same pixel may have different semantics in different regions of interest, and determined object category while generating object’s mask. Pham et al. came up with Biseg \cite{p11}. They used semantic segmentation fractional graph and Li et al.'s inside/outside position sensitive score maps as prior information, regarded instance masks as posterior information, and deduced object masks from prior information using Bayesian model. Wang et al. proposed SOLO \cite{p12}. This algorithm divided the input image into S×S grids, used FPN \cite{p13} to distinguish objects of different scales, and tried to directly segment masks from the image.
\begin{figure*}[!hbt]
\centering
\includegraphics[width=1.0\linewidth]{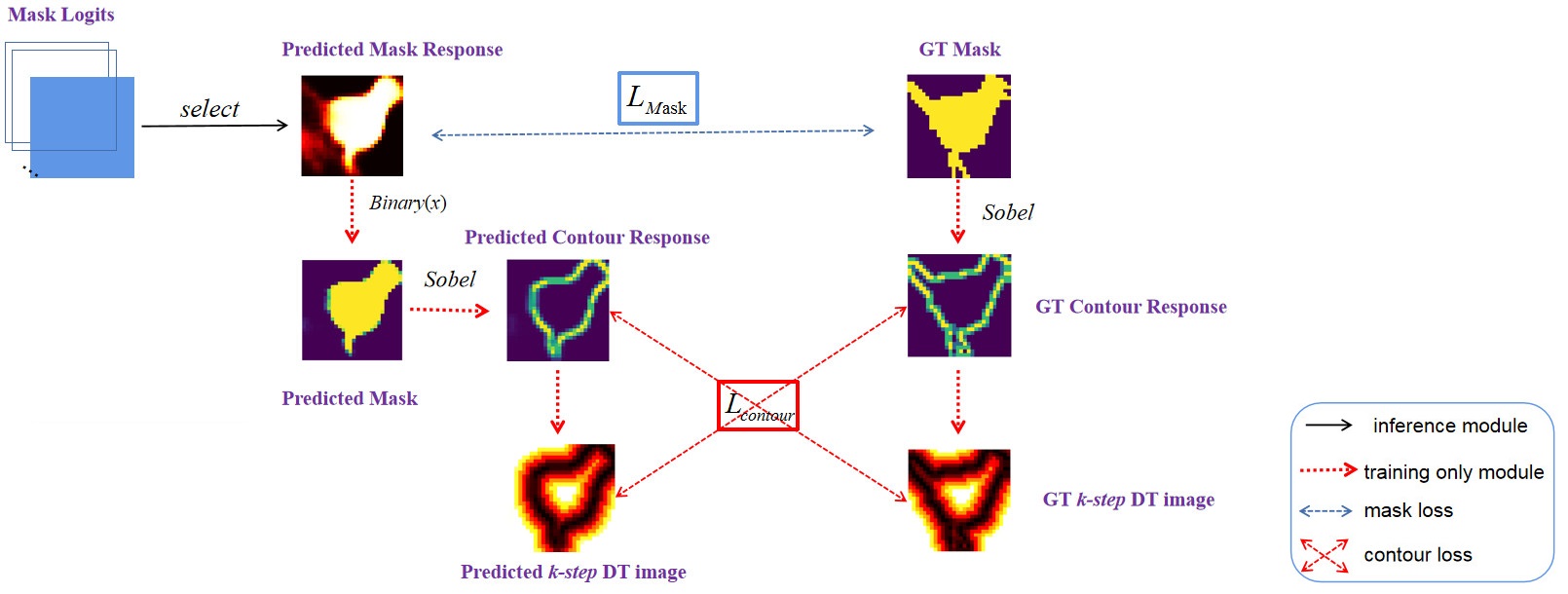}
\caption{\textit{The overall sketch of the proposed contour loss.}}
\label{fig1}
\end{figure*}

\emph{Detection-based} methods firstly rely on object detector to locate targets in the image, and then perform pixel-level classification within each target area. He et al. proposed Mask R-CNN in 2017 \cite{p2}, which took full advantage of object detector to achieve high instance segmentation accuracy. Since then, detection-based instance segmentation methods represented by Mask R-CNN have gradually become the mainstream. The essence of Mask R-CNN is to add a feature alignment module and a mask branch to Faster R-CNN \cite{p14}. Inspired by previous works, Fu et al. developed RetinaMask \cite{p15} which was a real-time single-stage instance segmentation algorithm. PA-Net \cite{p16} established information flow between low-level features and high-level features, which further improved the instance segmentation precision. Mask Scoring R-CNN \cite{p17} depended on a Mask IoU branch to handle the problem of mismatch between mask quality and mask score. HTC \cite{p18} made full use of reciprocal relationship between detection task and segmentation task to integrate and learn complementary features of each stage. Combining rich context information between the mask branches in different stages, it greatly improved instance segmentation accuracy. In general, the performance of detection-based methods is better than that of segmentation-based methods. Thus, we choose to evaluate our proposed method on Mask R-CNN framework.
\subsection{Methods Combining Edge or Boundary Information}\label{subsec2.2}
Recently, there are some attempts to incorporate edges or boundaries to facilitate instance segmentation.

Kang et al. \cite{p19} extended the edge of ground-truth mask to inside and outside by \emph{k} pixels, and assigned pixel values of the extended parts empirically. This method was conducive to learning richer edge information and achieved a little performance improvement in both object detection and instance segmentation. However, the method is very sensitive to hyper-parameter \emph{k}, which needs to be adjusted for different databases. Moreover, most values of \emph{k} contribute to negative gains. Instead, our method is built upon classic image distance transformation technology and has a solid theoretical foundation. It is not sensitive to hyper-parameter \emph{k}, and there is no hyper-parameter for loss fusion, which shows strong generality.

Roland et al. \cite{p20} used classical \emph{Sobel} \cite{p21} operator to extract edge images of predicted mask and ground-truth mask respectively. The error between edge images was measured by mean square error (MSE) loss, which improved instance segmentation accuracy of object’s edge. Beside edge images which only contain simple position information, we design \emph{k-step} DTI module to encode additional distance information, which can be essentially regarded as an active contour model and can learn the object’s contour better.

Hayder et al. \cite{p22} took DTI as a mask representation, and predicted DTI of ground-truth mask through a complex neural network branch. This method relayed on an explicit encode-decode module and special post-processing steps to produce objects’ masks. Although image distance transformation is good at describing the closeness of similar contour points, distance transformation values of regions far from the contour are easily affected by various disturbances. Therefore, the algorithm’s stability needs to be improved. Differently, we design a truncated DTI module which is inferable and differentiable. By truncation, the algorithm pays more attention to the optimization of the contour points. Inferable and differentiable characteristics make our truncated DTI to be used as an evaluation metric. When applying truncated DTI, the inference network structure of the original algorithm can be inherited and preserved to produce more accurate masks and no further post processing steps are needed.

Cheng et al. \cite{p23} trained a new branch to predict the edges of masks to exploit edge information, directly increasing training parameters. Our method does not need to modify network structure of basic algorithm and does not increase training parameters, only optimizing existing parameters.

In short, the main difference from the above works is that we design an inferable and differentiable implementation of truncated DTI, which can generate new supervisory information online to specifically optimize object’s contour part. Another difference is that we propose contour loss on the foundation of the truncated DTI, which achieves better performances compared with existing methods.
\section{Method}\label{sec3}
In this section, we firstly introduce the overall sketch of the proposed contour loss for instance segmentation. Then, we demonstrate the procedure of computing \emph{k-step} DTI, i.e. the truncated DTI, which is used for the computation of contour loss. In the end, we detail the mathematical definition as well as pseudocode of contour loss.
\subsection{Overall Architecture}\label{subsec3.1}
Mask R-CNN is a general instance segmentation framework, but it takes no consideration of the segmentation quality near the contour. To overcome this drawback, we design a contour loss function on the foundation of \emph{k-step} DTI and integrate it into Mask R-CNN to achieve joint training. The calculation process of contour loss is shown in Fig. \ref{fig1} Contour loss does not change the original network structure and can also be applied to other instance segmentation frameworks.

As shown in Fig. \ref{fig1}, the calculation process of contour loss starts from the mask branch’s output of the present instance segmentation method. Firstly, according to the prediction of the classification branch, the predicted mask response is selected from the mask branch. Secondly, a simulated binarization operation is conducted on the selected mask response to approximately obtain the predicted mask. Thirdly, a fixed parameter convolution layer with \emph{Sobel} operator as its convolution kernel is utilized to convolve the predicted mask and the ground-truth mask to get the predicted contour response and the ground-truth contour response respectively. Finally, image distance transformation operation is conducted on the predicted contour response and the ground-truth contour response to get the predicted \emph{k-step} DTI and the ground-truth \emph{k-step} DTI respectively. The coverage values of one contour response image onto \emph{k-step} DTI of the other contour response image are accumulated. Contour loss is defined as average of the two normalized coverage values. It can be jointly trained with the original mask loss to make the object mask more accurate and clearer near the contour. Various parts of the proposed contour loss are detailed as follows.
\begin{figure*}[!t]
\centering
\includegraphics[width=1.0\linewidth]{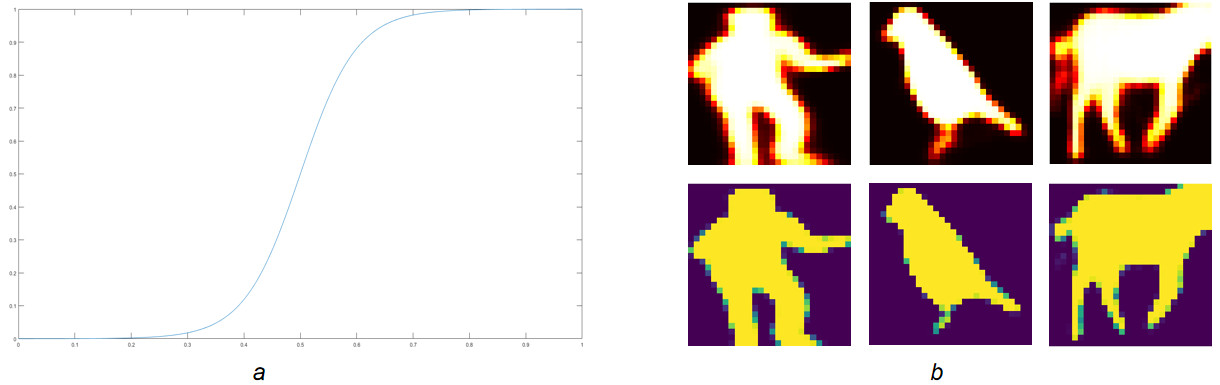}
\caption{\textit{The mathematical function curve for binarization (a) and the binarized results (b).}}
\label{fig2}
\end{figure*}
\begin{figure*}[!t]
\centering
\includegraphics[width=1.0\linewidth]{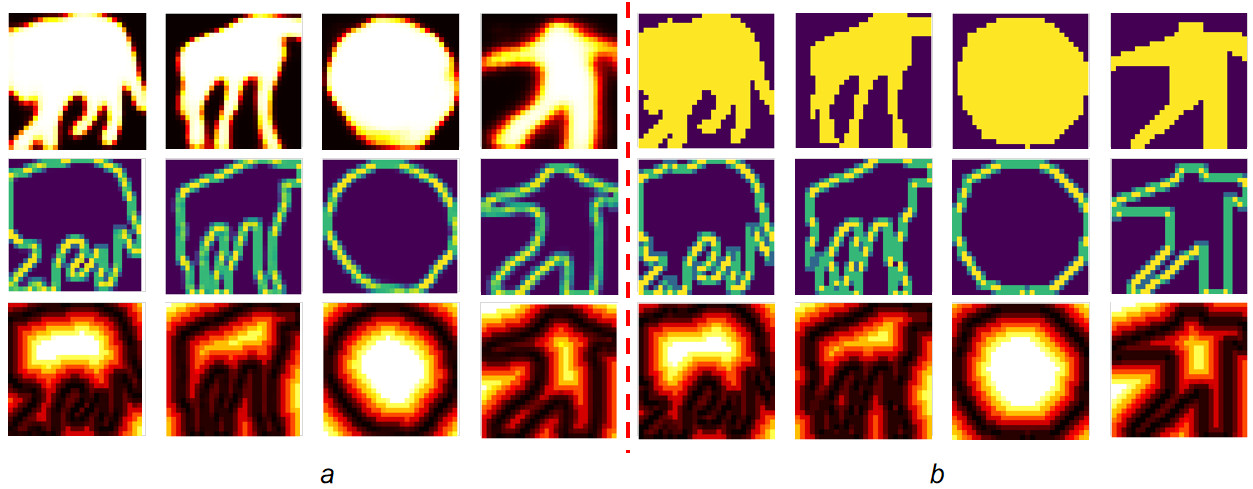}
\caption{\textit{Typical predicted mask responses (a) and ground-truth masks (b) are shown at the top row. Their corresponding contour responses and k-step DTIs are displayed at the middle row and bottom row, respectively.}}
\label{fig3}
\end{figure*}

\textbf{Binarization of the predicted mask response.} Denote $M_R$ be the predicted mask response selected from the output of the mask branch. We utilize a differentiable mathematical function $B(.)$ to approximately binarize it to obtain the predicted mask $M_P$.
\begin{align}
    B(x) & =\frac{1}{1+e^{-\gamma(x-T)}} \label{eq1} \\
    M_{P} & =B\left(M_{R}\right) \label{eq2}
\end{align}
where $\gamma$ and $T$ represent the slope and threshold (binarization value), respectively. We set $\gamma$ to 20 and $T$ to 0.5 by default. The purpose of using the mathematical function is to simulate the binarization operation in a differentiable manner, which corresponds to the step of obtaining the predicted binary masks at inference stage. The curve of the mathematical function is shown on Fig. \ref{fig2}(a). Several pairs of the predicted mask responses (first row) and their corresponding simulated binary images (second row) are shown on Fig. \ref{fig2}(b).

\textbf{Calculation of the contour response.} We construct a fixed parameter convolution layer with \emph{Sobel} operator as its convolution kernel to convolve the predicted mask $M_P$ and the ground-truth mask $M_{GT}$ in both $x$ and $y$ directions to obtain the predicted contour response $\Omega_{PCR}$ and the ground-truth contour response  $\Omega_{GCR}$, respectively.

\begin{equation}
    \emph{\text { Sobel }}_{x}=\left[\begin{array}{ccc}
    -1 & 0 & 1 \\
    -2 & 0 & 2 \\
    -1 & 0 & 1
    \end{array}\right], \emph{\text { Sobel }}_{y}=\left[\begin{array}{ccc}
    1 & 2 & 1 \\
    0 & 0 & 0 \\
    -1 & -2 & -1
    \end{array}\right]
    \label{eq3} \\
\end{equation}

\begin{align}
    \Omega_{PCR} & =\frac{1}{2}\left[|M_{P} * \emph{\text{ Sobel }}_{x}|+| M_{P} * \emph{\text { Sobel }}_{y} \mid\right] \label{eq4} \\
    \Omega_{GCR} & =\frac{1}{2}\left[|M_{GT} * \emph{\text{ Sobel }}_{x}|+| M_{GT} * \emph{\text { Sobel }}_{y} \mid\right] \label{eq5}
\end{align}
where $*$ is the standard convolution operation, and $|.|$ is the absolute value. Typical contour response images are shown in Fig. \ref{fig3}, where the first row of Fig. \ref{fig3}(a) and Fig. \ref{fig3}(b) respectively shows the predicted mask responses and the ground-truth masks. The second row of them shows their corresponding contour responses.

\textbf{DTI of the contour response.} The values of pixels in DTI that are far away from the contour may be unstable, thus can interfere with the optimization process. In order to make contour loss focus on optimizing the object’s contour parts, we use a threshold \emph{k} to truncate the DTI of the contour response. In other words, pixel values of DTI exceeding \emph{k} are set to \emph{k}, and the resulting image is called \emph{k-step} DTI. The last row of Fig. \ref{fig3} shows the computed \emph{k-step} DTIs. We can see that pixel values close to object’s contour are smaller (shown darker), while those far away from the contour are larger (shown brighter). The white areas indicate that their pixel values reach the truncated threshold \emph{k}.

Note that the computation of \emph{k-step} DTI must be differentiable, otherwise contour loss cannot be backpropagated during the network training phase. Therefore, we design an approximated differentiable module of the \emph{k-step} DTI under modern neural network structures, which is referred as \emph{kSDT} (\emph{k-Step} Distance Transformation) algorithm (see Subsection \ref{subsec3.2} for specific principle and implementation details). We apply \emph{kSDT} to the predicted contour response $\Omega_{PCR}$ and the ground-truth contour response $\Omega_{GCR}$ to compute the predicted \emph{k-step} DTI $\Gamma _{P}^{k}$ and the ground-truth \emph{k-step} DTI $\Gamma_{GT}^{k}$, respectively.
\begin{align}
    \Gamma_{P}^{k} & =kSDT\left(\Omega_{PCR}\right) \label{eq6} \\
    \Gamma_{GT}^{k} & =kSDT\left(\Omega_{GCR}\right) \label{eq7}
\end{align}

\textbf{Computation of the contour loss.} Each pixel value of \emph{k-step} DTI represents the distance between it and the closest point of the contour, which can be used to measure the difference between two contours. To compute contour loss, we firstly accumulate coverage values of one contour response image onto \emph{k-step} DTI of the other contour response image. Then, we regard the average of the two normalized coverage values as the difference measure between the predicted contour response and the ground-truth contour response. During training stage, when the predicted contour response deviates from the ground-truth contour response, contour loss will optimize and correct the predicted mask response, making the predicted mask obtained at inference stage more accurate and clearer near the contour. Specific principle and implementation details of contour loss are shown in Subsection \ref{subsec3.3}.

\textbf{Joint training with contour loss.} Numerous studies have shown that multi-task learning performs better than single-task learning. Thus, we define a multi-task loss for each training batch which is expressed as follows.
\begin{equation}
    L=L_{cls}+L_{ box}+L_{mask}+L_{Contour}
    \label{eq8}
\end{equation}
where the classification loss $L_{cls}$, box regression loss $L_{ box}$, and the mask loss $L_{ mask}$  are the same as those in Mask R-CNN. $L_{Contour}$ is the proposed contour loss (see Subsection \ref{subsec3.3}).
\begin{figure}[]
\includegraphics[width=1.0\linewidth]{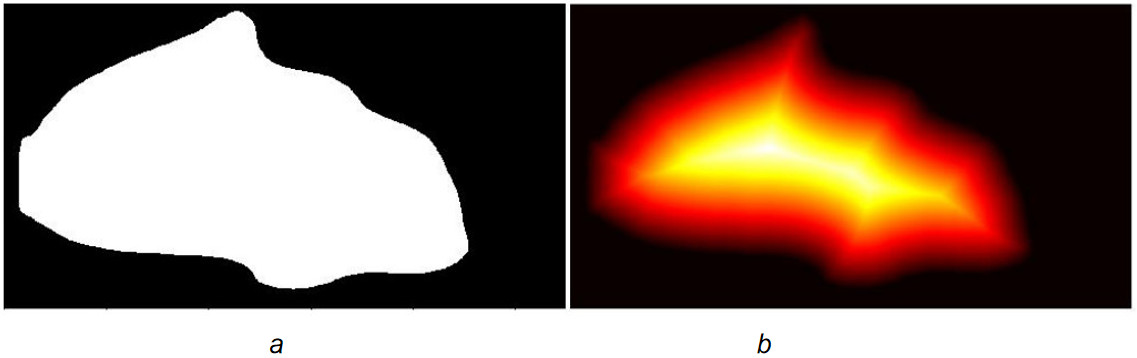}
\caption{\textit{A binary image(a) and its DTI(b) shown as a heat map.}}
\label{fig4}
\end{figure}
\begin{figure*}[]
\centering
\includegraphics[width=1.0\linewidth]{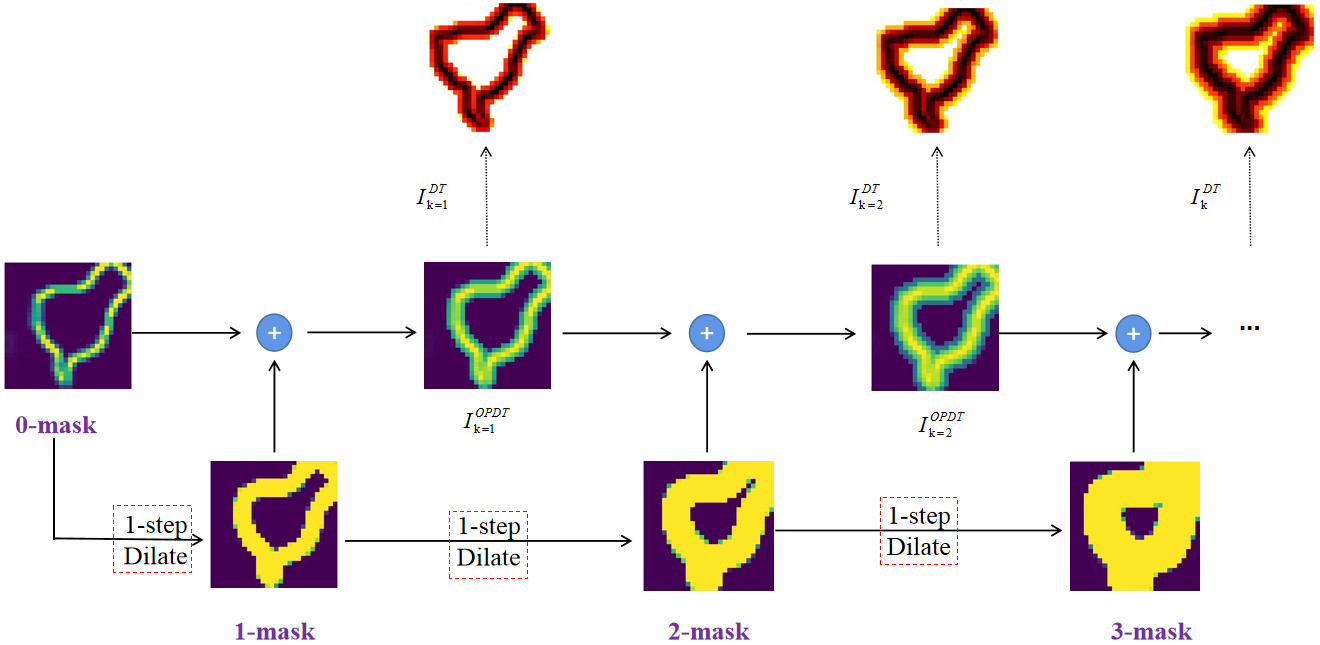}
\caption{\textit{The calculation flow chart of kSDT.}}
\label{fig5}
\end{figure*}
\subsection{k-Step DTI} \label{subsec3.2}
Before presenting the approximated differentiable implementation of the proposed \emph{k-step} DTI under modern neural networks, we first review the concept of image distance transformation. Image distance transformation is a classical technology in computer vision which has already been implemented in Open-CV, MATLAB and other common tools. DTI refers to a kind of gray image obtained after image distance transformation operation on an input binary image whose foreground pixel value is 1 and the background pixel value is 0. Each pixel value of DTI represents the distance between the pixel and the closest background pixel on the input binary image. Denote $b(i,j)$ be the pixel value of the input binary image at pixel $(i,j)$ and $t(i,j)$ be the pixel value of its DTI at pixel $(i,j)$. Naturally, $t(i,j)$ is 0 when $b(i,j)$ equals to 0, and $t(i,j)$ is greater than 0 when $b(i,j)$ equals to 1. In addition, $t(i,j)$ is small when its location is close to the background region of the binary image, while is large when its location is far away from the background region. Fig. \ref{fig4} shows a binary image (actually a binary mask of a car) and its DTI which is shown as a heat map. The brighter the pixel on the heat map, the larger its pixel value, and vice versa.

Different from the above common DTI, the calculation of \emph{k-step} DTI needs some minor changes. Given an initial binary contour image, points belonging to the contour are foreground and the rests are background. We then perform an opposite operation on it to make points belonging to the contour to be background, and the rests to be foreground. Finally, we apply image distance transformation operation to the obtained binary image with opposite value and use a truncation threshold \emph{k} to obtain the expected \emph{k-step} DTI. The acquired \emph{k-step} DTI describes the closest distance to the contour for each pixel, which can be effectively used to measure the difference between contours.

However, the computation of the above \emph{k-step} DTI has factors which are not differentiable, and there are no available modules under existing deep learning frameworks. To solve this problem, we propose an approximated differentiable implementation of the \emph{k-step} DTI suitable for current neural networks, called \emph{kSDT}. Fig. \ref{fig5} shows the calculation flow chart of \emph{kSDT}. The black solid arrow in Fig. \ref{fig5} represents the data flow, and the black dashed arrow represents the output of \emph{k-step} DTI.

The algorithm takes the contour response in Subsection \ref{subsec3.1} as initial input which is denoted as $I_{0}^{Mask}$ (\emph{0-mask}). By iteratively executing \emph{k} ($k \ge 1$) groups of \{$D(.)$, $\oplus$\} operations in formula (\ref{eq9}) and formula (\ref{eq10}), the \emph{k-step} DTI with opposite value is obtained which is denoted as $I_{k}^{OPDT}$. The final \emph{k-step} DTI can be calculated accordingly in formula (\ref{eq11}).
\begin{align}
    I_{k}^{Mask} & =D\left(I_{k-1}^{Mask }\right) \label{eq9} \\
    I_{k}^{OPDT} & =I_{k}^{Mask } \oplus I_{k-1}^{OPDT} \label{eq10} \\
    I_{k}^{D T} & =(k+1)-I_{k}^{OPDT} \label{eq11}
\end{align}
where $D(.)$ represents an \emph{one-step} dilation operator, and $\oplus$ represents element-wise addition. The above calculation process is differentiable except for the dilation operator $D(.)$.

\begin{figure}[]
\includegraphics[width=1.0\linewidth]{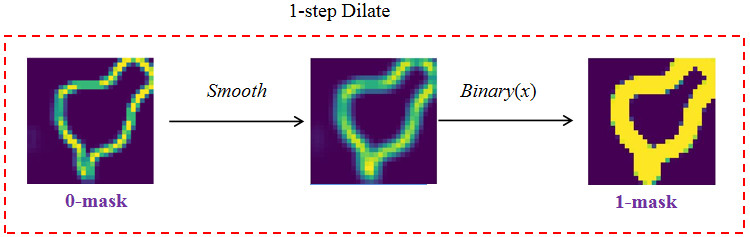}
\caption{\textit{Schematic diagram of the designed differentiable one-step dilation operator.}}
\label{fig6}
\end{figure}

To make the whole process differentiable, we further design an approximated differentiable \emph{one-step} dilation operator. Taking the computation of $I_{1}^{Mask}$ as an example, the specific calculation process is shown in Fig. \ref{fig6}. The algorithm firstly constructs a fixed parameter convolution layer with a \emph{smooth} operator in formula (\ref{eq12}) as its convolution kernel to convolve $I_{0}^{Mask}$ once. Then, formula (\ref{eq1}) is utilized to approximately binarize the smoothed image to get the expected dilated image $I_{1}^{Mask}$ (\emph{1-mask}). Here, we set $\gamma$ to 20 and $T$ to 0.1. By taking the dilated image as the input of the next stage, dilation result of each stage (\emph{k-mask}) can be iteratively obtained. Note that the input for \emph{one-step} dilation operator is not restricted to binary image, making the whole \emph{k-step} DTI module (\emph{kSDT}) compatible to \textbf{continuous} response maps. The calculation process of \emph{kSDT} is summarized in Algorithm \ref{alg1}.
\begin{equation}
\emph{\text { Smooth }}=\frac{1}{9}\left[\begin{array}{ccc}
1 & 1 & 1 \\
1 & 1 & 1 \\
1 & 1 & 1
\end{array}\right]
\label{eq12}
\end{equation}
\begin{algorithm}[]
	\caption{\emph{kSDT}}
	\label{alg1}
	\textbf{Input:} $I_{0}^{Mask}, k$ \\
	\textbf{Output:} $I_{k}^{DT}$
	\begin{algorithmic}[1]
		\STATE {$I_{k}^{DT}\gets0, I^{OPDT}\gets I_{0}^{Mask}, I^{Mask}\gets I_{0}^{Mask};$}
		\FOR{$i$ in $k$}
		\STATE {$I^{Mask}\gets D\left(I^{Mask}\right); \qquad \qquad\emph{//Equation(9)}$}
		\STATE {$I^{OPDT}\gets \left(I^{OPDT}+I^{Mask}\right); \emph{//Equation(10)}$}
		\ENDFOR
		\STATE {$I_{k}^{DT}\gets \left \{ \left ( k+1 \right ) - I^{OPDT} \right \}; \qquad \emph{//Equation(11)}$}
		\STATE {$\textbf{return} \quad I_{k}^{DT};$}
	\end{algorithmic}
\end{algorithm}
\begin{figure*}[]
\centering
\includegraphics[width=1.0\linewidth]{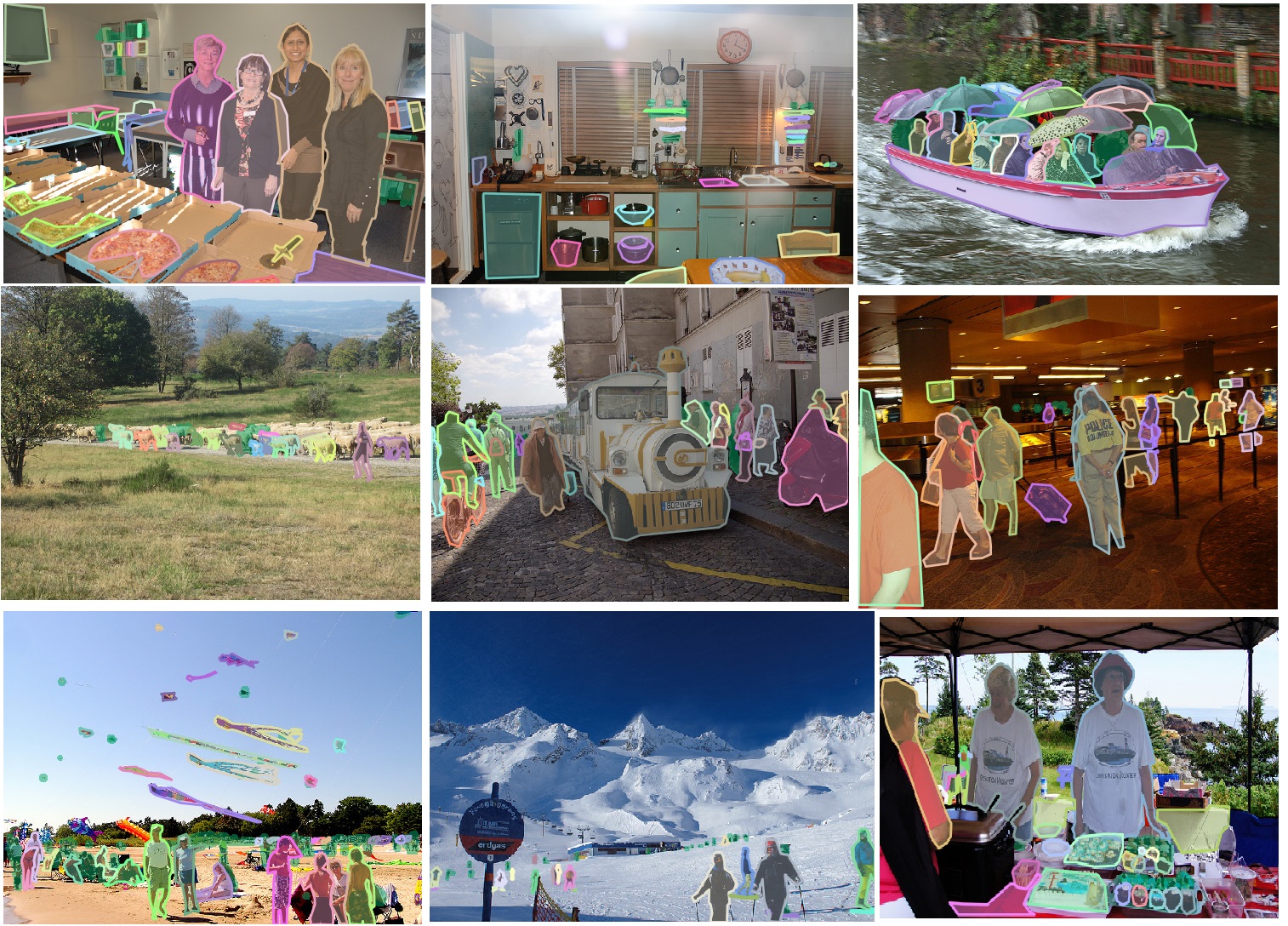}
\caption{\textit{Typical training images with polygon mask annotations in COCO 2014.}}
\label{fig7}
\end{figure*}
\subsection{Contour Loss Function} \label{subsec3.3}
Contour loss can be measured by the distance between the predicted contour and the ground-truth contour. Mathematically, let $\Omega_{GT}$ and $\Omega_{P}$ be the ground-truth contour and the predicted contour, respectively. $A$ represents a contour point on $\Omega_{P}$, then the distance between $A$ and $\Omega_{GT}$ is usually defined as the distance between $A$ and its closest ground-truth contour point:
\begin{equation}
d\left(A, \Omega_{GT}\right)=\min _{B \in \Omega_{GT}}\{d(A, B)\} \label{eq13}
\end{equation}

Let $\Gamma_{GT}$ and $\Gamma_{P}$ be DTIs of $\Omega_{GT}$ and $\Omega_{P}$ , respectively. According to the process of computing DTI, the distance between $A$ and $\Omega_{GT}$ in formula (\ref{eq13}) is equal to the coverage value on $\Gamma_{GT}$ by $A$:
\begin{equation}
d\left(A, \Omega_{G T}\right)=\Gamma_{G T}(A) \label{eq14}
\end{equation}

Therefore, the distance between the predicted contour and the ground-truth contour can be computed based on DTIs:
\begin{align}
d\left(\Omega_{P}, \Omega_{G T}\right) & =  \frac{1}{2}\left[\frac{1}{\left|\Omega_{P}\right|} \sum_{A \in \Omega_{P}} \Gamma_{G T}(A) \nonumber \right.+\\
  &\phantom{=\;\;}\left.\qquad\frac{1}{\left|\Omega_{G T}\right|} \sum_{B \in \Omega_{G T}} \Gamma_{P}(B)\right] \label{eq15}
\end{align}
where $\left|\Omega_{\mathrm{GT}}\right|$ and $\left|\Omega_{\mathrm{P}}\right|$ are the numbers of contour points for $\Omega_{GT}$ and $\Omega_{P}$, respectively.

In order to ensure the differentiable property, we design a continous version of formula (\ref{eq15}) to calculate contour loss which employs the continous contour responses and their \emph{k-step} DTIs:
\begin{align}
d\left(\Omega_{PCR}, \Omega_{GCR}\right) &= \frac{1}{2}\left[\frac{GAP\left(\Omega_{PCR} \otimes \Gamma_{GT}^{k}\right)+\varepsilon}{GAP\left(\Omega_{PCR}\right)+\varepsilon} \nonumber \right.+\\
&\phantom{=\;\;}\left.\qquad \frac{GAP\left(\Omega_{GCR} \otimes \Gamma_{P}^{k}\right)+ \varepsilon}{GAP\left(\Omega_{GCR}\right)+\varepsilon}\right] \label{eq16}
\end{align}
where $\bigotimes$ represents Hadamard product, $\emph{\text{GAP}}(.)$ represents global average pooling, and $\varepsilon$ is a smooth term to avoid zero division.

Assuming that a total of $N$ positive samples are obtained in a training batch, the final contour loss function can be expressed as:
\begin{equation}
L_{Contour}=\frac{1}{N} \sum_{i=1}^{N} d\left(\Omega_{P C R}^{i}, \Omega_{G C R}^{i}\right) \label{eq17}
\end{equation}

Algorithm \ref{alg2} provides detailed calculation process of contour loss, where $M_{PR}$ represents the predicted mask response, $M_{P}$ and $M_{GT}$ represent the predicted mask and the ground-truth mask respectively, $\emph{\text{ConvSobel}}(.)$ represents the convolution operation with $\emph{\text{Sobel}}$ kernel, and $L_{Contour}$ represents contour loss.
\begin{algorithm}[]
    \setstretch{0.6}
	\caption{Calculation Process of $L_{Contour}$}
	\label{alg2}
	\textbf{Input:} \text{Images from COCO} \\
	\textbf{Output:} $L_{Contour}$
	\begin{algorithmic}[1]
		\STATE {$L_{Contour}\gets0;$}
		\FOR{image-annotation $\left \{ I, M_{GT} \right \}$ in COCO}
		\STATE {$MaskLogits\gets MaskR$-$CNN(I);$}
		\FOR{$\left\{M_{PR,i},M_{GT,i} \right \}$ in a batch $\left \{MaskLogits,M_{GT} \right \}$}
		\STATE {$M_{PR,i}\gets \emph{\text{select}}\left( M_{PR,i}\right);$}
		\STATE {$M_{P,i}\gets B\left(M_{PR,i} \right);\quad \qquad\qquad\qquad\quad\emph{//Equation(2)}$}
        \STATE {$\Omega_{PCR,i}\gets ConvSobel\left( M_{P,i}\right);\;\;\;\qquad\ \ \ \, \emph{//Equation(4)}$}
		\STATE {$\Omega_{GCR,i}\gets ConvSobel\left( M_{GT,i}\right);\quad\qquad\emph{//Equation(5)}$}
		\STATE {$Calculate$ $\Gamma_{P,i}, \Gamma_{GT,i};\ \quad\qquad\qquad\qquad\emph{//\textbf{Algorithm \ref{alg1}}}$}
		\STATE {$Calculate$ $L_{Contour}\left(\Omega_{PCR,i},\Omega_{GCR,i} \right);\emph{//Equation(16)}$}
		\ENDFOR
		\STATE {$L_{Contour}\gets \left(L_{Contour} + L_{Contour}\left(\Omega_{PCR,i},\Omega_{GCR,i} \right) \right);$}
		\ENDFOR
		\STATE {$L_{Contour}\gets L_{Contour}/N ;$}
	\end{algorithmic}
\end{algorithm}
\section{Experiments}\label{sec4}
\subsection{Dataset and Metrics} \label{subsec4.1}
In order to verify the effectiveness and the generalization ability of the proposed contour loss, we conducted extensive experiments on COCO which is a widely used benchmark dataset for common objects instance segmentation. This dataset is challenging due to the large number of target categories and the wide ranges of object scales. Fig. \ref{fig7} shows some typical training images with polygon mask annotations. We use COCO 2014 to do experiments which includes 82783 training images and 40504 validation images. We train models on the whole training set, and report results on the mini validation set which contains 5000 images. We use the standard metrics, i.e. COCO AP, to evaluate all models including: mAP, AP50, AP75, APs, APm, APl.
\begin{table*}[ht]
\caption{The impact of different values of \emph{k} on mask accuracy($\%$). $L_{Contour}$ represents contour loss.\label{tab1}}
\begin{tabular*}{\textwidth}{@{\extracolsep{\fill}}lccccccc}\toprule
Method &mAP & AP50  & AP75 & APs & APm & APl & \emph{k}\\
\midrule
baseline &34.28 &55.94 & 36.20 & 15.82 & 36.75 & 50.84 & $-$ \\
$L_{Contour}$ &34.39 &55.88 &36.46 & 15.81 & 36.98 & 51.26 & 1  \\
$L_{Contour}$ &34.54 &56.16 &36.54 & 16.26 & 36.95 & 51.42 & 2 \\
$L_{Contour}$ &34.39 &56.10 &36.19 & 15.74 & 36.74 & 51.18 & 3 \\
$L_{Contour}$ &34.36 &55.89 &36.42 & 15.82 & 36.76 & 51.13 & 4 \\
$L_{Contour}$ &34.36 &56.02 &36.37 & 15.99 & 36.79 & 50.99 & 5 \\
$L_{Contour}$ &34.18 &55.91 &36.13 & 15.92 & 36.50 & 50.74 & 6 \\
\bottomrule
\end{tabular*}{}
\end{table*}
\begin{figure*}[]
\includegraphics[width=1.0\linewidth]{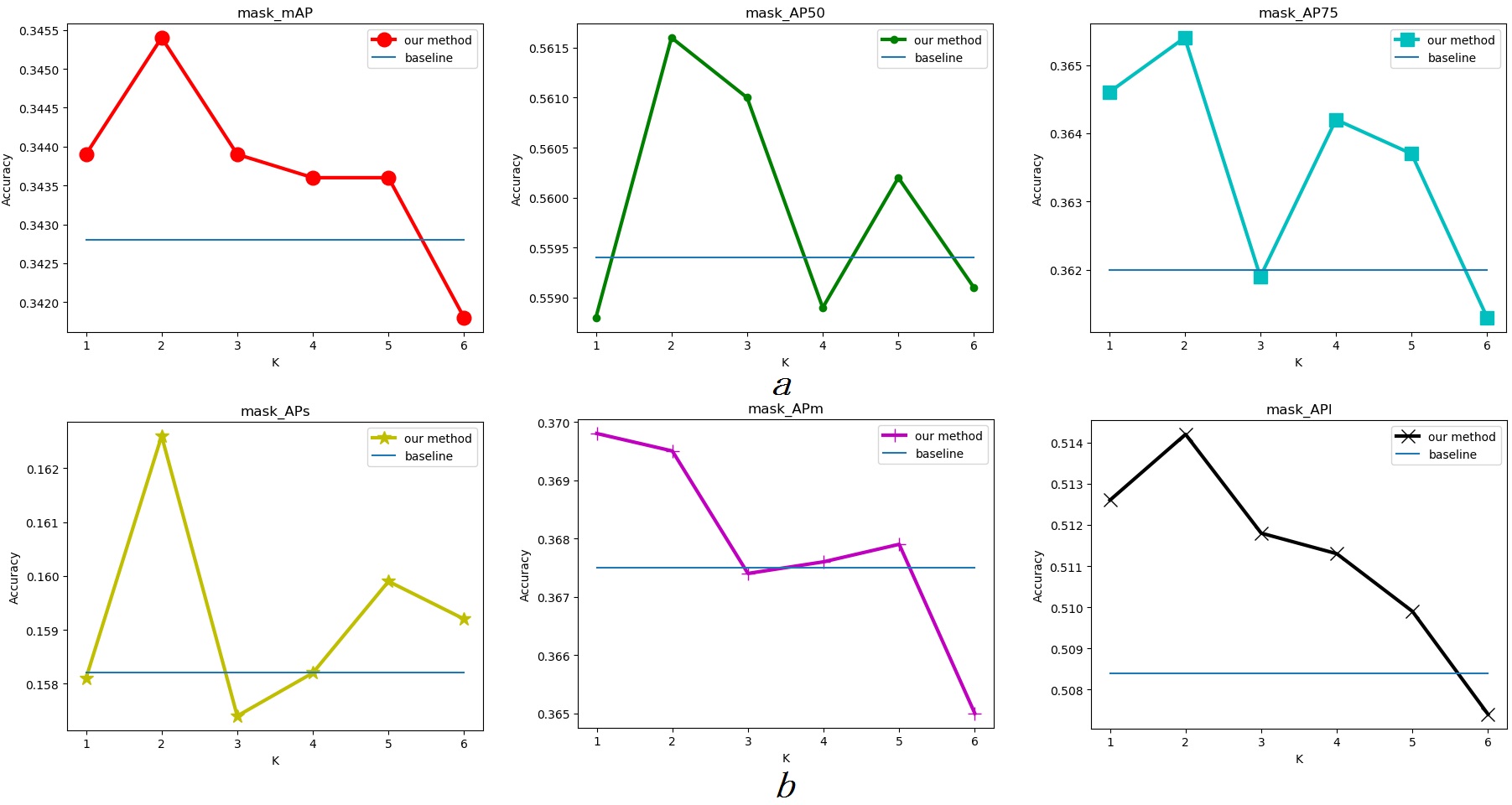}
\caption{\textit{Visualization of the data of Table 1. (a): mAP, AP50, AP75. (b): APs, APm, APl.}}
\label{fig8}
\end{figure*}
\subsection{Implementation Details and Experimental Setup} \label{subsec4.2}
We implement contour loss on top of Mask R-CNN for its efficiency and good performance. We use Res-50+FPN \cite{p24} as the backbone network by default. Our code is based on the open source project mask r-cnn benchmark \cite{p25}. We initialize the backbone network with the weights pre-trained on Image-Net \cite{p26}. We train the instance segmentation network for a total of 180K iterations. We set the initial learning rate to 0.01 and reduce it by a factor of 0.1 and 0.01 after 120K and 160K iterations respectively. We train all models on 4 NVIDIA 2080Ti GPUs utilizing SGD with 8 images per mini-batch. Unless specified, the input image is resized to have 800 pixels along the shorter side and their longer side less than or equal to 1333. Other hyper-parameters are kept consistent with the open source project. For larger backbones, we follow the linear scaling rule \cite{p27} to adjust the learning rate schedule when decreasing mini-batch size.

It’s noteworthy that contour loss usually plays an auxiliary role. In other words, we only enable contour loss after some iterations of the original algorithm. Specifically, we firstly train the original Mask R-CNN to 120K iterations (save as a checkpoint), then enable the contour loss (load the saved checkpoint), and continue to train it to 180K iterations. Naturally, our baseline is the Mask R-CNN trained from 120K (load the saved checkpoint) to 180K iterations without contour loss. On one hand, this can reduce the verification time of the proposed method. On the other hand, it may prevent instability of the mask branch caused by contour loss at early training stage.
\begin{table*}[ht]
\caption{Performance comparison of different loss functions($\%$).\label{tab2}}
\begin{tabular*}{\textwidth}{@{\extracolsep{\fill}}lcccccc}\toprule
Method &mAP & AP50  & AP75 & APs & APm & APl \\
\midrule
baseline &34.28 &55.94 & 36.20 & 15.82 & 36.75 & 50.84 \\
\\
$L_{Edge}^{M S E}$ &34.33 &55.87 &36.39 & 15.79 & 36.87 & \underline{51.02} \\
\\
$L_{Contour}^{M S E}$ &\underline{34.42} &\underline{55.95} &\textbf{36.55} & \underline{15.88} & \textbf{37.01} & 51.00 \\
\\
$L_{Contour}$ &\textbf{34.54} &\textbf{56.16} &\underline{36.54} & \textbf{16.26} & \underline{36.95} & \textbf{51.42} \\
\bottomrule
\end{tabular*}{\\ \\ The highest value in each column is shown in bold, and the second highest value is underlined.}
\end{table*}
\begin{table*}[ht]
\caption{Performance comparison of different instance segmentation algorithms($\%$). MR represents Mask R$-$CNN. CL represents contour loss.\label{tab3}}
\begin{tabular*}{\textwidth}{@{\extracolsep{\fill}}ccccccccc}\toprule
Method & Backbone    & \multicolumn{1}{c}{CL} & mAP   & AP50  & AP75  & APs   & APm   & APl   \\
\midrule
MR     & Res-50+FPN  & $\times$                      & 34.28 & 55.94 & 36.20 & 15.82 & 36.75 & 50.84 \\
MR     & Res-50+FPN  & $\surd$                       & 34.54 & 56.16 & 36.54 & 16.26 & 36.95 & 51.42 \\
MR     & Res-101+FPN & $\times$                      & 35.79 & 58.02 & 38.16 & 16.75 & 38.70 & 53.09 \\
MR     & Res-101+FPN & $\surd$                       & 35.96 & 58.04 & 38.35 & 16.52 & 38.89 & 53.50 \\
MR     & Res-X-101   & $\times$                      & 38.16 & 61.00 & 40.93 & 18.61 & 40.91 & 55.27 \\
MR     & Res-X-101   & $\surd$                       & 38.29 & 61.16 & 41.19 & 18.54 & 41.01 & 55.67 \\
HTC*    & Res-50+FPN  & $\times$                      & 37.7  & 59.1  & 40.2  & 19.3  & 40.4  & 53.4  \\
HTC*    & Res-50+FPN  & $\surd$                       & 37.9  & 59.3  & 40.3  & 19.2  & 40.6  & 53.1  \\
\bottomrule
\end{tabular*}{\\ \\ *Note that the implementation of HTC is based on mm-detection \cite{p28}. We train it on COCO \emph{2017 train} (115K images) and report results on COCO \emph{2017val} (5K images).}
\end{table*}
\subsection{Evaluation on Hyper-parameter k} \label{subsec4.3}
An important parameter involved in calculation of contour loss is \emph{k}. We explored the impact of different values of \emph{k} on mask accuracy. We selected 6 different values of \emph{k} for experiments, including: 1,2,3,4,5,6. The results are summarized in Table \ref{tab1}. From the table we can see that \emph{k} is not sensitive for contour loss in the ranges of 1$\sim$5. Under the auxiliary supervision of contour loss, most of the evaluation metrics of the benchmark algorithm have been improved to a certain extent. Enjoying contour loss can achieve the maximum gain of 0.26\% mAP, 0.22\% AP50, 0.34\% AP75, 0.44\% APs, 0.2\% APm, and 0.58\% APl respectively. We set \emph{k} to 2 in the following experiments.
\begin{figure*}[]
\centering
\includegraphics[width=1.0\linewidth]{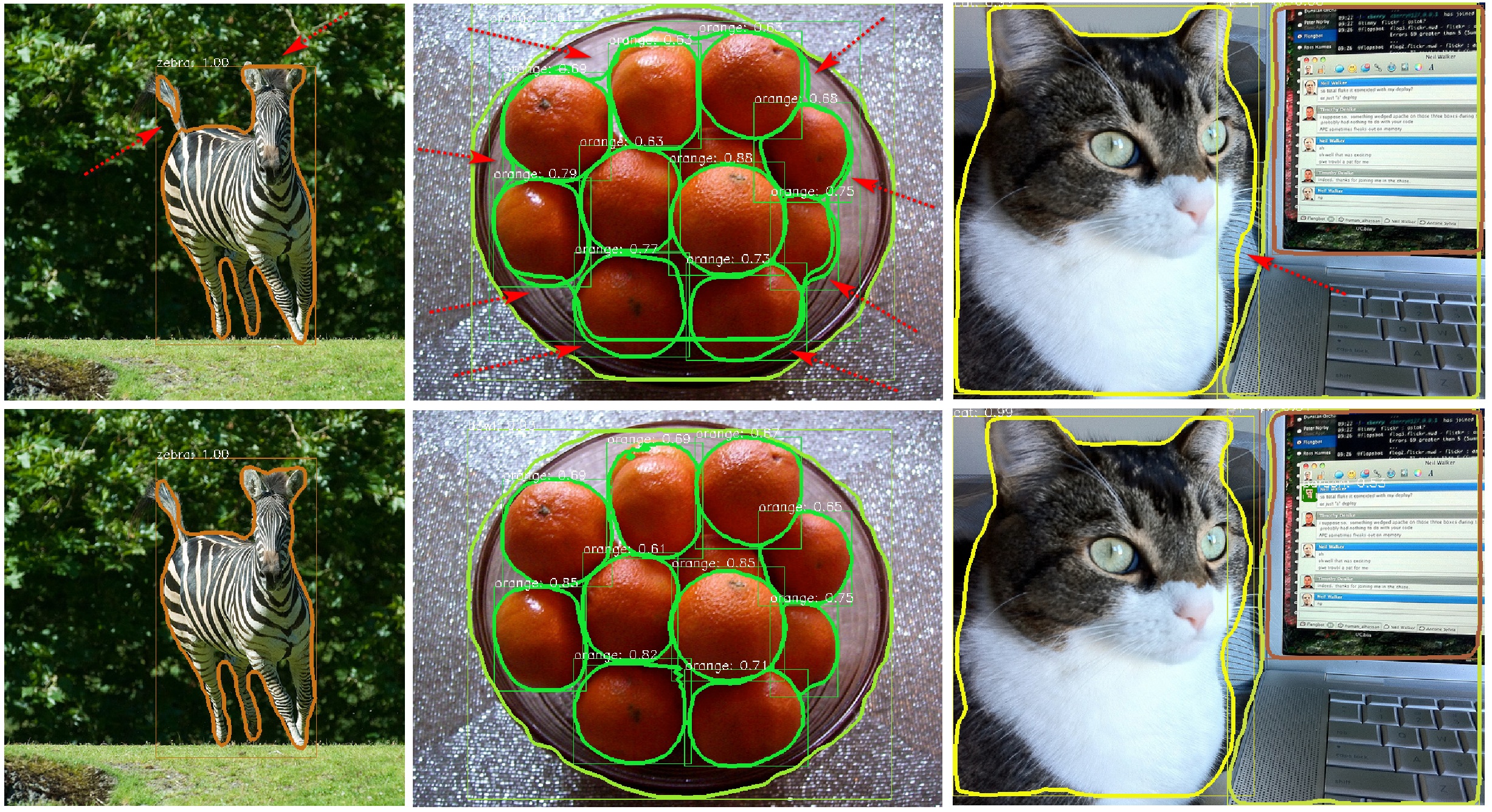}
\caption{\textit{Qualitative comparison between the benchmark algorithm (top row) and our method (bottom row).}}
\label{fig9}
\end{figure*}

Fig. \ref{fig8} visualizes the data of Table \ref{tab1}. The horizontal axis represents the value of \emph{k}, and the vertical axis represents mask accuracy (represented by decimals). Each subplot represents the comparison result under one kind of metric. It can be seen from the figures that under the auxiliary supervision of contour loss, mask accuracy can be improved on most metrics.
\subsection{Ablation Study} \label{subsec4.4}
In order to further verify the performance of contour loss, we choose two alternatives to it for ablation experiments.
\begin{itemize}
    \item MSE Edge Loss: We utilize MSE loss to calculate the distance between the predicted contour response $\Omega_{PCR}$ and the ground-truth contour response $\Omega_{GCR}$.
    \begin{equation}
        L_{Edge}^{M S E}=\frac{1}{N} \sum_{i=1}^{N}\left(\Omega_{P C R}^{\mathrm{i}}-\Omega_{G C R}^{i}\right)^{2} \label{eq18}
    \end{equation}
where $N$ is the total number of positive samples.
    \item MSE Contour Loss: We utilize MSE loss to reduce the error between the predicted \emph{k-step} DTI $\Gamma_{P}^{k}$ and the ground-truth \emph{k-step} DTI $\Gamma_{GT}^{k}$.
    \begin{equation}
        L_{Contour}^{M S E}=\frac{1}{N} \sum_{i=1}^{N}\left(\Gamma_{P, i}^{k}-\Gamma_{G T, i}^{k}\right)^{2} \label{eq19}
    \end{equation}
    where $N$ is the total number of positive samples.
\end{itemize}

Experimental results are summarized in Table \ref{tab2}. The last line represents the proposed contour loss. From the table we can observe that: (1) Compared with the benchmark algorithm, all of the three loss functions can improve the mask accuracy. (2) Contour loss is superior to the other two loss functions, and it can get the best mask accuracy.
\subsection{Comparative Study} \label{subsec4.5}
In order to verify the generalization ability of contour loss, we choose to conduct comparative experiments on Mask R-CNN with different backbones and HTC with Res-50+FPN backbone. Experimental results are summarized in Table \ref{tab3}. The proposed contour loss respectively brings the gains of 0.13\%$\sim$0.26\% mAP, 0.16\%$\sim$0.22\% AP50, 0.26\%$\sim$0.34\% AP75, 0.44\% APs, 0.1\%$\sim$0.2\% APm and 0.4\%$\sim$0.58\% APl on Mask R-CNN. HTC with contour loss achieves the gains of 0.2\% mAP, 0.2\% AP50, 0.1\% AP75 and 0.2\% APm respectively. Thus, contour loss is effective for different instance segmentation methods.
\subsection{Qualitative Analysis} \label{subsec4.6}
Fig. \ref{fig9} shows the qualitative segmentation results of Mask R-CNN (top row) and “Mask R-CNN + Contour Loss” (bottom row). The results are based on backbone network of Res-50+FPN. For the convenience of comparison, we only show the contours of the predicted masks. By comparing the areas indicated by red dashed arrow, we can see that object masks segmented by our method have more accurate and clearer contours, which proves the effectiveness of the proposed method.
\section{Conclusions}\label{sec5}
In this paper, we introduce classic distance transformation image (DTI) into instance segmentation. We propose a contour loss function based on the designed differentiable \emph{k-step} DTIs to specifically optimize the contour parts of the predicted masks. Contour loss can be effectively integrated into existing instance segmentation methods and combined with their original loss functions to gain more accurate and clearer masks. The proposed method does not need to modify the original network structure or increase more training parameters, thus has strong versatility. Experiments on COCO show that contour loss is effective and can further improve the performance of current instance segmentation methods. In future work, we will explore the possibility of applying contour loss to instance segmentation of unseen objects.
\section*{Acknowledgment}
This work is partly supported by National Natural Science Foundation of China (Grant No. U19B2033, Grant No.62076020), and National Key R\&D Program (Grant No. 2019YFF0301801).



\end{document}